\documentclass[11pt,a4paper]{article}
\usepackage[hyperref]{acl2018}
\usepackage{times}
\usepackage{latexsym}
\usepackage{footnote}
\usepackage{array,booktabs,makecell,tablefootnote}
\usepackage{longtable}
\usepackage{float}
\restylefloat{table}

\usepackage[utf8]{inputenc}
\usepackage[arabic,USenglish]{babel}
\usepackage{enumitem}

\usepackage{booktabs}
\usepackage{todo} 
\usepackage{url}

\usepackage{color}
\usepackage{graphicx}
\usepackage[export]{adjustbox}
\usepackage{enumitem}
\usepackage{array}
\usepackage{float}
\usepackage{bbding}
\usepackage{multirow}

\aclfinalcopy 

\title{Named Entity Recognition on Code-Switched Data:\\
Overview of the CALCS 2018 Shared Task}

\author{Gustavo Aguilar $^\ddagger$, Fahad AlGhamdi, Victor Soto $^\dagger$, \\
{\bf Mona Diab, Julia Hirschberg, $^\dagger$\and Thamar Solorio$^\ddagger$ } \\ 
 Department of Computer Science, The George Washington University \\
{\tt \{fghamdi, mtdiab\}@gwu.edu} \\
$^\ddagger${Department of Computer Science, University of Houston} \\
{\tt $^\ddagger${\{gaguilaralas, tsolorio\}@uh.edu}}\\
$^\dagger${Department of Computer Science, Columbia University} \\
{\tt $^\dagger${\{vs2411, julia\}@cs.columbia.edu}}}

\date{}

\begin{document}
\maketitle

\begin{abstract}
In the third shared task of the Computational Approaches to Linguistic Code-Switching (CALCS) workshop, we focus on Named Entity Recognition (NER) on code-switched social-media data. We divide the shared task into two competitions based on the English-Spanish (ENG-SPA) and Modern Standard Arabic-Egyptian (MSA-EGY) language pairs. We use Twitter data and 9 entity types to establish a new dataset for code-switched NER benchmarks. In addition to the CS phenomenon, the diversity of the entities and the social media challenges make the task considerably hard to process. As a result, the best scores of the competitions are 63.76\% and 71.61\% for ENG-SPA and MSA-EGY, respectively. We present the scores of 9 participants and discuss the most common challenges among submissions. 
\end{abstract}

\section{Introduction}
Code-switching (CS) is a linguistic behavior that occurs on spoken and written language. CS happens when multilingual speakers move back and forth from one language to another in the same discourse. The growing incidence of social media in the way we communicate has also increased the occurrences of code-switching on informal written language. As a result, there is a prevalent demand for more tools and resources that can help to process such phenomenon.

In the previous versions of the Computational Approaches to Linguistic Code-Switching (CALCS) workshop, we focused on providing an annotated corpora for language identification \citep{solorio-EtAl:2014:CodeSwitch, molina-EtAl:2016:W16-58}. In this occasion, we extend the annotations to the Named Entity Recognition (NER) level. The goal of this shared task is to provide a code-switched NER dataset that can help to benchmark NER state-of-the-art approaches. This will directly impact the performance of higher-level NLP applications where the code-switching behavior is commonly found.

\begin{figure}[h]
\renewcommand{\arraystretch}{1.3}
\centering
\small
\begin{tabular}{cc}
\begin{tabular}{|l|}
\hline
{\bf ENG-SPA Tweet } \\\hline
\underline{\textbf{Original:}} @\_xoxoBecky lmao ni ganas tengo de llorar \\
\includegraphics[width=1em]{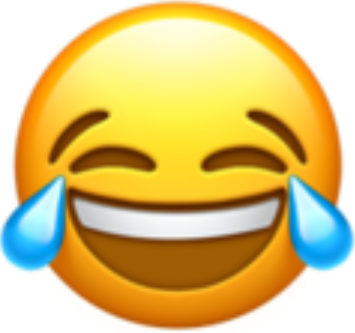} , the last movie that made me cry
was [\textbf{\textit{Pineapple}} \\
\textbf{\textit{Express}}]\textsubscript{TITLE} \includegraphics[width=1em]{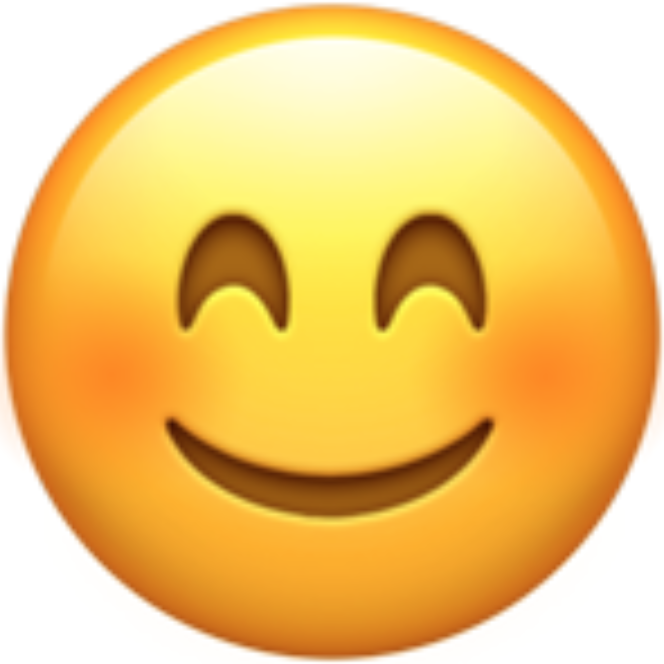} me dejo llorando de risa 
\includegraphics[width=1em]{images/emoji1.png}\includegraphics[width=1em]{images/emoji1.png} \\
\underline{\textbf{English:}} @\_xoxoBecky lmao I don't even want to cry \\
\includegraphics[width=1em]{images/emoji1.png} , the last movie that made me cry was [\textbf{\textit{Pineapple}} \\ 
\textbf{\textit{Express}}]\textsubscript{TITLE} \includegraphics[width=1em]{images/emoji2.png} it left me crying with laughter \includegraphics[width=1em]{images/emoji1.png}\includegraphics[width=1em]{images/emoji1.png} \\
\hline
\end{tabular} 
\\\\

\begin{tabular}{|l|l|}
\hline
{\bf MSA-EGY Tweet  } \\\hline
\underline{\textbf{Buckwalter Encoding:}} wAy mErkp Dd [\textbf{\textit{AldAxlyp}}]\textsubscript{ORG} \\

[\textbf{\textit{wAmn Aldwlp}}]\textsubscript{ORG} hbqY sEydp byhA  \\
\underline{\textbf{Arabic:}} \textRL{ وأي معركة ضد الداخلية وأمن الدولة} \\

\textRL{ هبقى سعيدة بيها}
\\
\underline{\textbf{English:}} Any controversy against the Interior Ministry  \\
and State Security Service will make me feel happy  \\
\hline
\end{tabular}
\end{tabular}
\caption{ Examples of the CALCS 2018 dataset. In the English-Spanish data, the highlighted words represent a movie, tagged as TITLE. While in the MSA-EGY data, the bolded words represent government agencies, tagged as ORGANIZATION }
\label{f:example}
\end{figure}
%


We had a total of 9 participants from which we received 8 submissions on English-Spanish and 5 submissions on Modern Standard Arabic-Egyptian. The best F1-score reported for ENG-SPA\footnote{ENG-SPA competition~\url{https://competitions.codalab.org/competitions/18725}} was \textbf{63.76\%} by the \textbf{IIT BHU} team \citep{IITBHU:2018:CALCS} whereas in MSA-EGY\footnote{MSA-EGY competition~\url{https://competitions.codalab.org/competitions/18724}} 
was \textbf{71.61\%} by the \textbf{FAIR} team \citep{FAIR:2018:CALCS}. 

\section{Task definition}

The task consists of recognizing entities in a relatively short code-switched context. The entity types for this task are \textit{person}, \textit{organization}, \textit{location}, \textit{group}, \textit{title}, \textit{product}, \textit{event}, \textit{time}, and \textit{other}. We describe each entity type on Section \ref{entity_defs}. Since NER is a sequential tagging task, we use the IOB scheme to identify multiple words as a single named entity. The addition of this scheme duplicates the number of entities in the task yielding a B(eginning) and I(nside) variations of each of them. This leaves us with 19 possible labels for the classification task. 

The evaluation of the task uses two versions of the F1-score. The first is the standard F1, and the second is the Surface Form F1-score introduced by \citet{DBLP:journals/corr/DerczynskiM0EGTPB14}. The Surface Form F1-score captures the rare and emerging aspects of the entities. We average both metrics to determine the positions in the leaderboard. Additionally, the shared task was conducted on the CodaLab platform\footnote{The competitions will be permanently open for future benchmarks}, where participants are able to directly evaluate their approaches against the gold data.

\section{Datasets}

In this section we provide the definition of our labels, describe the annotation process and show the distribution of the ENG-SPA and MSA-EGY datasets.

\begin{table*}[t!]
\centering
\setlength{\tabcolsep}{13pt}
\begin{tabular}{lllllll}
\toprule
\multirow{2}{*}{\textbf{Classes}} &
	\multicolumn{3}{c}{\textbf{ENG-SPA} } &
	\multicolumn{3}{c}{\textbf{MSA-EGY} } \\
				& \textbf{Train}	& \textbf{Dev}  & \textbf{Test}		
                & \textbf{Train} 	& \textbf{Dev}	& \textbf{Test}\\\midrule
Person 			& 6,226		& 95	& 1,888 	& 8,897	 	& 1,113	  	& 777 \\
Location 		& 4,323		& 16	& 803   	& 4,500	 	& 474	  	& 332 \\
Organization	& 1,381		& 10	& 307   	& 2,596	 	& 263	  	& 179 \\
Group	     	& 1,024		& 5		& 153   	& 2,646	 	& 303	  	& 139 \\
Title			& 1,980		& 50	& 542   	& 2,057	 	& 258	  	& 18 \\
Product			& 1,885		& 21	& 481   	& 795	 	& 81	  	& 54 \\
Event			& 557		& 6		& 99    	& 902	 	& 121	  	& 81 \\
Time			& 786		& 9		& 197   	& 578	 	& 79	  	& 28 \\
Other			& 382		& 7		& 62    	& 122	 	& 19	  	& 2 \\\hline
NE Tokens		& 18,544	& 219 	& 4,532		& 23,093	& 2,711	  	& 1,610 \\
O Tokens 		& 614,013	& 9,364	& 178,479 	& 181,229	& 20,031	& 19,804 \\\hline
Tweets 			& 50,757	& 832	& 15,634	& 10,102	 		& 1,122	  		& 1,110 \\\bottomrule
\end{tabular} 
\caption{The named entity distribution of the training, development and testing sets for both language pairs. Note that the \textit{NE tokens} row contains the B(eginning) and I(nside) tokens of the datasets following the IOB scheme. The \textit{O Tokens} row refers to the non-entity tokens. }
\label{t:data_distr}
\end{table*}

\subsection{Entity instructions} 
\label{entity_defs}
The named entities have been annotated using the instructions below. Note that the definitions of the entity types apply to both language pairs.

\begin{itemize}
\item \textbf{Person}: This entity type includes proper names and nicknames that can
identify a person uniquely. We ignore cases where a person is referred by nouns with
adjectives that are not necessarily a nickname. Single artists and famous people are
treated as \textit{person}. 

\item \textbf{Organization}: This entity type includes names of companies, institutions
and corporations, i.e. every entity that has employees and takes actions as a whole.
If the NE can potentially be any other type, the context should be sufficient to support whether it is \textit{organization} or not (e.g., Facebook as organization vs. Facebook as the website application).

\item \textbf{Location}: This NE refers to physical places that people can visit. It 
includes cities, countries, addresses, facilities, touristic places, etc. This entity type is not to be confused with \textit{organization}. For instance, when people use organization names to refer to places that can be visited (e.g., restaurants), those entities must be tagged as \textit{location}.

\item \textbf{Group}: This NE includes sports teams, music bands, duets, etc. \textit{Group} and \textit{organization} are not to be confused. For example, the Houston Astros as a team (i.e., \textit{group}) is different from the Houston Astros institution. 

\item \textbf{Product}: This NE refers to articles that have been manufactured or
refined for sale, like devices, medicine, food produced by a company, any well-defined
service, website accounts, etc. 

\item \textbf{Title}: This type includes titles of movies, books, TV shows, songs,
etc. Very often, titles can be sentences (e.g., the movie \textit{We're the Millers}).
\textit{Titles} usually refer to media and must not be confused with the \textit{product} type.

\item \textbf{Event}: This type refers to situations or scenarios that gather people
for a specific purpose such as concerts, competitions, conferences, award events, etc.
\textit{Events} do not consider holidays.

\item \textbf{Time}: This NE includes months, days of the week, seasons, holidays and
dates that happen periodically, which are not \textit{events} (e.g., Christmas). It excludes hours, minutes, and seconds. `Yesterday', `tomorrow', `week' and `year' are not tagged as \textit{time}.

\item \textbf{Other}: This type includes any other named entity that does not fit in the previous categories. This may include nationalities, languages, music genres, etc.
\end{itemize}

The motivation behind these entity types partly lies on the contextual difference in which they appear. For instance, when an \textit{organization} can be lexically confused with a \textit{product}, the context should break down the ambiguity. Additionally, we tried to
include entity types that have an impact on higher-level NLP applications under similar
social media scenarios.

\subsection{ENG-SPA}

\begin{figure}[t]
\centering
\includegraphics[width=\linewidth]{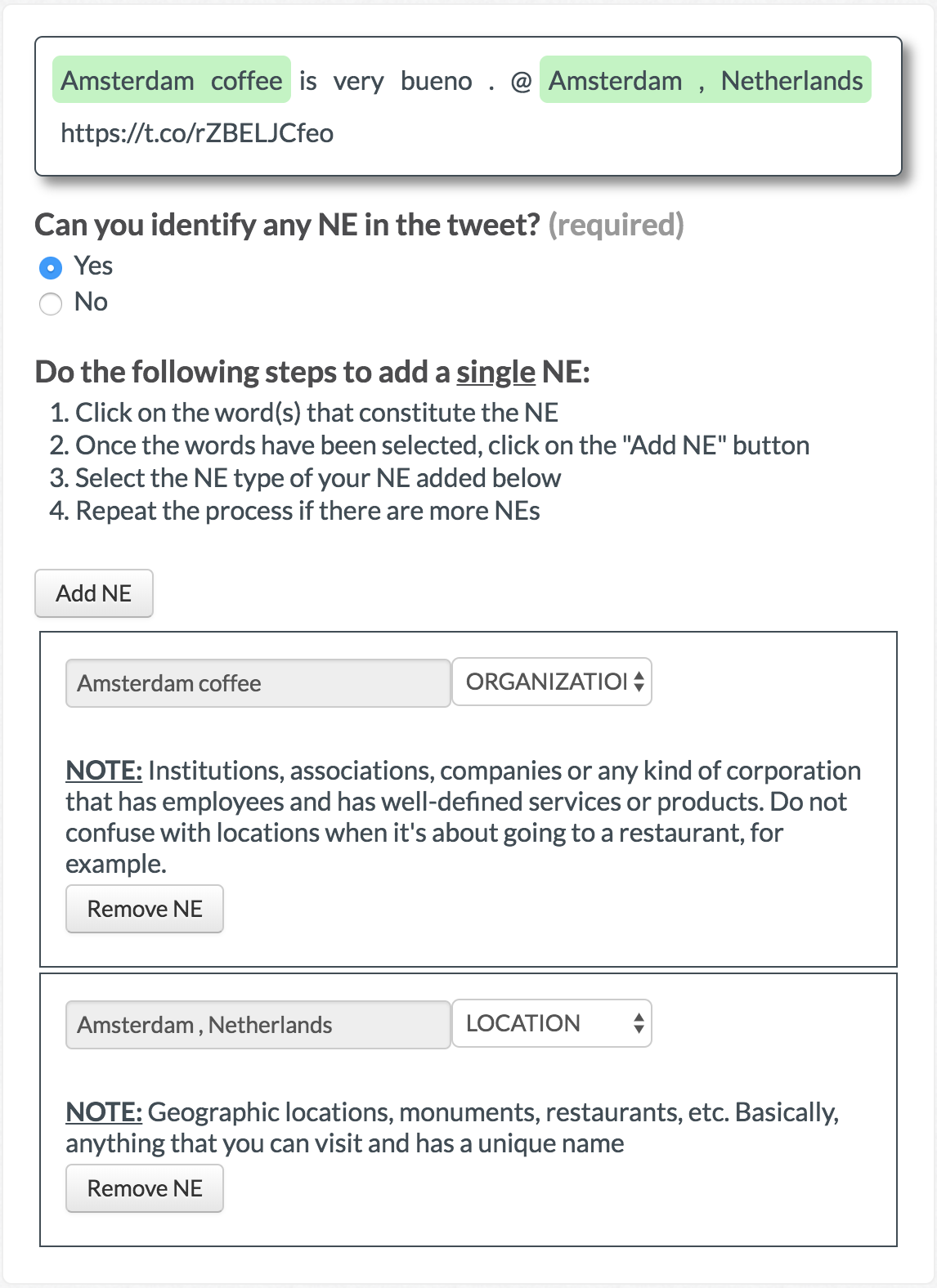}
\caption{The CrowdFlower interface that we developed to annotate the ENG-SPA dataset.
	 The green-highlighted words are the entities selected by the annotator.
	 The words in the same green area describe a single entity.
	 Once the NE selection has been added, the annotators have to select the
	 type of the entities. 
}
\label{f:eng-spa-crowdflower}
\end{figure}

\noindent \textbf{Data annotation}: We use the English-Spanish language identification dataset introduced in the first CALCS shared task \citep{solorio-EtAl:2014:CodeSwitch}. We build upon this dataset to generate the entity labels. To annotate the data, we designed a CrowdFlower\footnote{\url{https://crowdflower.com/}} job from scratch\footnote{The JavaScript code and HTML/CSS can be found here:~\url{https://github.com/tavo91/ner_annot}}. The interface of the job is described in Figure \ref{f:eng-spa-crowdflower}. The job allows annotators to select one or many words for a single NE. When the annotators select a word the tool suggests to incorporate words surrounding the current selection. When the selection of a whole entity is done, the annotators can add the entity to the second step where the type is determined. The annotators repeat this process until no more named entities can be identified in the tweet. The output of our customized job contains the entity type of one or multiple words that identify an NE according to the criteria of the annotators. The annotators are required to know both English and Spanish, and the job is constrained to reach an accuracy of at least 80\%. We also required 3 annotators per tweet. Additionally, the job was launched in geographic locations were both English and Spanish are reasonably common. Some of these places were USA, Mexico, Central America, Puerto Rico, Colombia, Venezuela, Chile, Uruguay, Paraguay and Spain. After getting the output data from CrowdFlower, we reviewed the results to correct any possible mistakes.

\noindent \textbf{Data distribution}: The entity types along with their distribution are listed in Table \ref{t:data_distr}. We provide training, development and testing\footnote{We do not provide the annotations of the test set because we want the CodaLab competition to be used for public benchmark in the future} sets containing 50,757, 832 and 15,634 tweets, respectively. The development and testing splits are inherited from previous CALCS Shared Tasks, whereas training uses the original split with the addition of 40,000 tweets. We added more tweets to the original training set to increase the number of samples per entity type since the NER datasets are naturally skewed. From Table \ref{t:data_distr}, it is worth noting that the total number of NE training tokens is 18,544 whereas the non-entity tokens add up to 614,013. This means that only 3\% of the tokens of the training set are NE-related. Likewise, the ratio of tokens for the development and testing sets are 2.3\% and 2.5\%, respectively. This skewed distribution poses a great challenge considering that the datasets are further separated by 18 fine-grained entity types (i.e., each entity type has a \textit{beginning} and \textit{inside} variations from the IOB scheme). However, we think that the skewness can be reasonably handled with the provided data. Moreover, the training, development and testing sets draw a very similar data distribution, which can also help to adapt the learning from training to testing.

\subsection{MSA-EGY}
\noindent \textbf{Validating old tweets}: For the Modern Standard Arabic-Egyptian Arabic Dialect (MSA-EGY) language pair, we combined the training, development, and test sets that we used in the EMNLP 2016 CS Shared Task~\cite{molina-EtAl:2016:W16-58} to create the new training corpora for the NER Shared Task. The data was harvested from Twitter. We apply a number of quality and validation checks to insure the quality of the old data.
Therefore, we retrieved all old tweets using the the new version of the Arabic Tweets Token
Assigner which is made available through the Shared Task website \footnote{\url{https://code-switching.github.io/2018/}}. One of the main reasons for the re-crawling step is to eliminate the tweets that have been deleted, or the tweets that belong to the users whose accounts are suspended by Twitter. The other reason is that some tweets may cause encoding issues when they are retrieved  using the crawler script. Thus, all these tweets were removed and eliminated. After performing the validation checks, we accepted and published 11,224 tweets (10,102 tweets for the training set, and 1,122 tweets for the development set). 

\textcolor{white}{.}

\noindent \textbf{Data creation and annotation}: Since we combined the test set used in the EMNLP-2016 CS Shared Task \cite{molina-EtAl:2016:W16-58} with the dataset used in the EMNLP-2014 CS Shared Task \cite{solorio-EtAl:2014:CodeSwitch} to form the new training and development sets, we needed to crawl and annotate a new test set for our new Shared Task. We resorted to using the Tweepy library to harvest the timeline of 12 Egyptian public figures. We applied the same filtration criteria when crawling and building the test set used in the 2016 CS shared task \cite{molina-EtAl:2016:W16-58}. We divided the old combined tweets into training and development sets as follows: 80\% train set and 10\% development set. Thus, we needed $\sim$ 1,110 tweets, which represents the 10\% of the new test set. As we did in the previous Shared Task, we wanted to consider choosing tweets from public figures whose tweets contain more code-switching points. Therefore, we resorted to using the Automatic Identification of Dialectal Arabic (AIDA2) tool \cite{AIDA2} to perform token-level language identification for the MSA and EGY tokens in context. Public figures with more than 35\% of code-switching points in their tweets were considered. The annotation work of the MSA-EGY dataset was done in-lab by two trained Egyptian native speakers. Our annotation team followed the Named Entity Annotation Guidelines for MSA-EGY, which is made available through the Shared Task website \footnote{\url{https://code switching.github.io/2018/}}. In the two previous editions of the CS Shared Task \cite{solorio-EtAl:2014:CodeSwitch,molina-EtAl:2016:W16-58}, we used a Named Entity (``ne'') tag. The ``ne'' tag was defined as a word or multi-word that represents names of a unique entity such as people’s names, countries and places, organizations, companies, websites, etc. The AIDA2 tool \cite{AIDA2} was used to assign initial automatic tags for highly confident data categories (i.e., URL, Punctuation, Number, etc) in addition to named entities. Then, we extracted and prepared all the tweets that contained ``ne" for annotation. As we mentioned earlier, the IOB scheme is used as an annotation scheme to identify multiple words as a single named entity. All the URLs, Punctuation and Numbers tags are deterministically converted to ``O'' tag, while the tweets that include ``ne'' tags were given to our in-lab annotators for validation and re-annotation if needed. 

\begin{figure}[t]
\centering
\includegraphics[width=.75\linewidth]{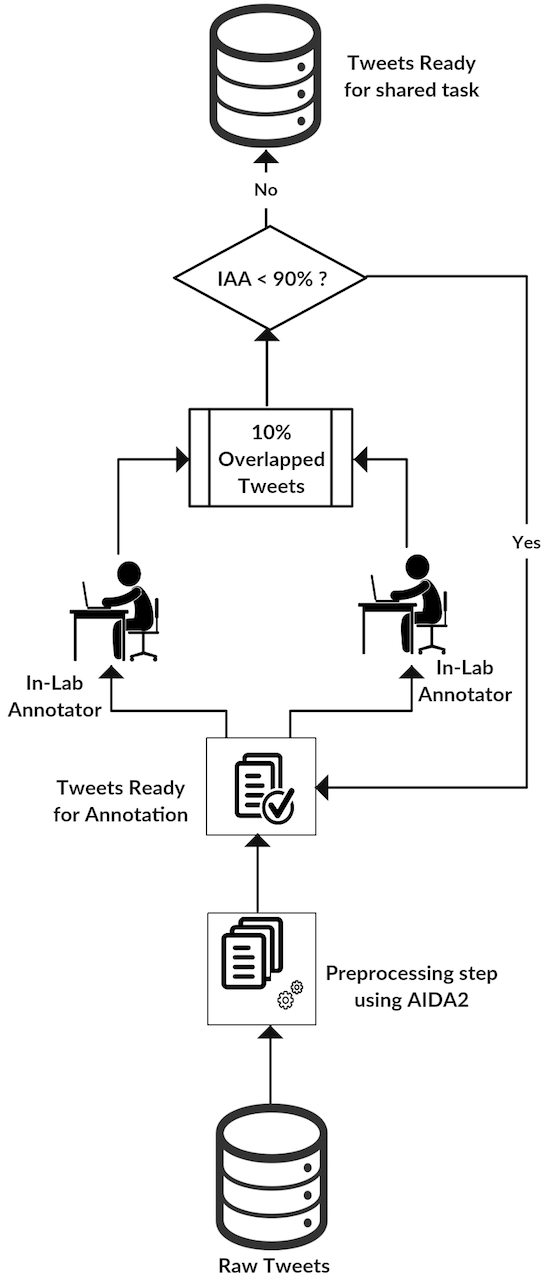}
\caption{MSA-EGY Data Annotation}
\label{f:msa-egy-annotation-workflow}
\end{figure}

\textcolor{white}{.}

\noindent\textbf{Quality checks and data distribution}: We computed the Inter-Annotator Agreement (IAA) on 10\% of the dataset to validate the performance and agreement among annotators. One of our annotators is a specialist linguist who carried out adjudication and revisions of accuracy measurements. We approached a stable Inter Annotator Agreement (IAA) of over 92\% pairwise agreement. The workflow of the annotation process for MSA-EGY is shown in Figure-\ref{f:msa-egy-annotation-workflow}. 

The total number of tweets in MSA-EGY dataset is 12,334 tweets. It is divided into three sets train, development, and test sets (10,102, 1,122, 1,110 tweets, respectively). Table \ref{t:data_distr} shows that the total number of NE training tokens is 23,093. It means that NE tokens represent 11.3\% of the total number of tokens. Similarly, the percentages of NE tokens in the development and test sets are 7.5\%, 11.9\%, respectively. As we mentioned earlier, the MSA-EGY tweets were harvested from the timeline of 12 Egyptian politicians public figures. Generally, politicians tend to use NEs more often when they write their tweets. This explains why the percentage of the NE tokens in MSA-EGY dataset is higher than the percentage of the NE tokens in ESP-ENG dataset.

\begin{table*}[t!]
\centering
\renewcommand{\arraystretch}{1.1}
\centering
\begin{tabular}{lccccccc}
\toprule
\textbf{Team}	& \textbf{Preproc}	& \textbf{Ext Res}	& \textbf{Hand Feats}	& \textbf{CNN}	 & \textbf{B-LSTM} & \textbf{CRF}	 & \textbf{Other} \\
\midrule 
IIT BHU 		&  	       	 & \checkmark &  			& \checkmark & \checkmark 	& \checkmark & MTL \\
CAiRE++			&  	       	 & 		 	  &  			&  			 & \checkmark	&  			 & FastText \\
FAIR			& \checkmark & 		 	  &  			&  			 & \checkmark	& \checkmark & Attention \\
Linguists 		&  	       	 & \checkmark & \checkmark 	&  			 &  			& \checkmark & \\
Flytxt	 		&  	       	 & \checkmark &  			&  			 &  			& \checkmark & \\
semantic		&  	       	 & 		 	  &  			&  			 & \checkmark	& \checkmark & \\
BATs			&  	       	 & \checkmark & \checkmark	&  			 &  			& \checkmark & \\
Fraunhofer FKIE	&  	       	 & \checkmark & \checkmark	&  			 &  			&  			 & SVM \\
GHHT			&  	       	 & \checkmark &  			&  			 & \checkmark	& \checkmark & \\
\bottomrule
\end{tabular}
\caption{The table shows the main component and strategies used by the participants. Ext Res means external resources such as pre-trained word embeddings, gazetteers, etc. Hand Feats means handcrafted features such as capitalization.}
\label{t:system_characteristics}
\end{table*}

\section{Approaches}
In this section, we briefly describe the systems of the participants and discuss their results as well as the final scores.

\begin{itemize}
\item\textbf{IIT BHU}~\citep{IITBHU:2018:CALCS}. They proposed a ``new architecture based on gating of character- and word-based representation of a token''. They captured the character and the word representations using a CNN and a bidirectional LSTM, respectively. They also used the Multi-Task Learning on the output layer and transfer the learning to a CRF classifier following \citet{aguilar-EtAl:2017:WNUT}. Moreover, they fed a gazetteers representation to their model. 

\item \textbf{CAiRE++}~\citep{CAiRE++:2018:CALCS}. They used a bidirectional LSTM model for characters and words. They primarily focused on OOV using the FastText library \citep{bojanowski2016enriching}. 

\item \textbf{FAIR}~\citep{FAIR:2018:CALCS}. They proposed a joint bidirectional LSTM-CRF network that uses attention at the embedding layer. They also preprocessed the data before feeding the network. 

\item\textbf{Linguists}~\citep{Linguists:2018:CALCS}. They used a Conditional Random Fields with many handcrafted features. Their focus was primarily on English-Spanish data.

\item\textbf{Flytxt}~\citep{Flytxt:2018:CALCS}. This team also employed a Conditional Random Fields. They fed the CRF with features from both external and internal resources. Additionally, they incorporated the language identification labels of the datasets from the previous versions of this workshop. 

\item\textbf{semantic}~\citep{semantic:2018:CALCS}. They jointly trained a Bidirectional LSTM with a Conditional Random Fields on the output layer. 

\item\textbf{BATs}~\citep{BATs:2018:CALCS}. They used a Conditional Random Fields with multiple features. Some of those features were also used for neural network, but they got better results with the CRF approach.

\item\textbf{Fraunhofer FKIE}~\citep{FraunhoferFKIE:2018:CALCS}. They used a Support Vector Machine (SVM) classifier with a Radial Basis kernel.  They handcrafted a lot of features and also included gazetteers.
		
\item\textbf{GHHT}~\citep{GHHT:2018:CALCS}. They trained a BLSTM-CRF network using pre-trained word embeddings, brown clusters and gazetteers. 

\item\textbf{Baseline}. We used a simple Bidirectional LSTM network with randomly initialized embedding vectors of 200 dimensions. We also used dropout operations on each direction of the BLSTM component. 
\end{itemize}

\section{Evaluation and results}

\subsection{Evaluation}

The evaluation of the shared task was conducted through CodaLab, where the participants were able to obtain immediate feedback of their submissions. The metrics used for the evaluation phase were the standard harmonic mean F1-score and the Surface Form F1 variation proposed by \citet{DBLP:journals/corr/DerczynskiM0EGTPB14}. Additionally, to have a single leaderboard per language pair, we unified both metrics by averaging them. The average values are the ones described in Table \ref{t:ranking}.

As stated by \cite{DBLP:journals/corr/DerczynskiM0EGTPB14}, the idea of the Surface Form F1-score is to capture the \textit{novel} and \textit{emerging} aspects that are usually encountered in social media data. Those aspects describe a fast-moving language that constantly produces new entities challenging more the recall capabilities of state-of-the-art models than the precision side.

\begin{table}[t]
\centering
\renewcommand{\arraystretch}{1.1}
\begin{tabular}{ll}
\toprule
\textbf{Team}				& \textbf{ENG-SPA} \\\midrule
\textbf{IIT BHU}	& \textbf{63.7628} \\
CAiRE++				& 62.7608 \\
FAIR				& 62.6671 \\
Linguists			& 62.1307 \\
Flytxt				& 59.2501 \\
semantic			& 56.7205 \\
BATs				& 54.1612 \\
Fraunhofer FKIE		& 53.6514 \\
Baseline			& 53.2802 \\\midrule
					& \textbf{MSA-EGY} \\\midrule
\textbf{FAIR}		& \textbf{71.6154} \\
GHHT 				& 70.0938 \\
Linguists			& 67.4419 \\
BATs				& 65.6207 \\
semantic			& 65.0276 \\
Baseline			& 62.7084 \\\bottomrule
\end{tabular}
\caption{The results of the participants in both ENG-SPA and MSA-EGY language pairs. The scores are based on the average of the standard and the Surface form F1 metrics. The highlighted teams are the best scores of the shared task.}
\label{t:ranking}
\end{table}

\subsection{Results and Error analysis}

Although all the scores reported by the participants outperformed the baselines in both ENG-SPA and MSA-EGY language pairs, the results are arguably low considering that the current state-of-the-art systems achieve around 91.2\% of F1-score on well-formatted text \citep{DBLP:journals/corr/LampleBSKD16,MaAndHovy:16,DBLP:journals/corr/abs-1709-04109}. As mentioned before, the best performing systems reached 63.76\% \citep{IITBHU:2018:CALCS} and 71.61\% ~\citep{FAIR:2018:CALCS} for ENG-SPA and MSA-EGY, respectively. These low outcomes are aligned with the challenges that come along with social media data and the addition of more heterogeneous entity types \citep{Ritter:2011:NER:2145432.2145595, AugensteinEtAl:17, DBLP:journals/corr/DerczynskiM0EGTPB14, aguilar-EtAl:2018:N18-1}.

Most of the MSA-EGY tweets are related to politics because they were harvested from the timeline of number of Egyptian politician public figures. Generally, these kinds of tweets encompass more NEs in comparison with other kinds of tweets. This explains why the percentage of the NE tokens in MSA-EGY dataset is high compared to the NEs' percentage in ESP-ENG data set. This high percentage of NE tokens helps the submitted systems to see and learn more examples and patterns. Thus, systems can generalize more effectively.

According to the results of the participants in the ENG-SPA shared task, the top three most challenging entity types were \textit{event}, \textit{title}, and \textit{time}. It is worth noting that these three classes are more or less the least frequent types in the dataset (see Table \ref{t:data_distr}), which suggests that having more data samples would produce better results. However, in the case of \textit{title}, there are 1,980 samples against 1,381 samples of \textit{organization}, and the performance is significantly better for the latter one (19\% vs. 35\% of F1-scores). Additionally, looking at Table \ref{t:challenging_test}, the entity \textit{Orange is the New Black} was not recognized by participants as a \textit{title}. This is an example of what we refer to heterogeneous entity type, meaning that the entity instances are flexible in format that can even describe independent sentences (i.e., a homogeneous type is \textit{person}). The entities \textit{Love Man} (title), \textit{Billboard 2014} (event), and \textit{show de shamu} (event) also describe the same pattern and they were hardly identified by participants.

\begin{table}
\renewcommand{\arraystretch}{1.1}
\begin{center}
\begin{tabular}{|l|l|} \hline 
\bf N & \bf ENG-SPA Samples \\ \hline
1 & Retiro totalmente lo dicho sobre \textbf{Orange} \\ 
  & \textbf{is the New Black}. Temporada terminada \\
  & y holly sh*t. HOLLY SH*T. \\\hline
2 & \textbf{Love Man} by \underline{\textbf{Otis Redding}}, found with \\
  & @Shazam. Listen now: como me hubiese\\
  & gustado ver a mis padres bailando esto ... \\\hline
3 & \underline{\textbf{Michael Jackson}} revivi\'o en los \\
  & \textbf{Billboard 2014} \\\hline
4 & @fairy0821 en el \textbf{show de shamu} !!! \\\hline
\end{tabular}
\end{center}
\caption{Challenging samples from the test set. The bold words are the ground truth samples and the underscored words are the predictions of the best performing systems.}
\label{t:challenging_test}
\end{table}

Unlike English and Spanish language pair which can be considered as two distinct languages, Modern Standard Arabic and Egyptian are more closely related which makes the task of identifying NE tokens more challenging. This is mainly due to the fact that Modern Standard Arabic and Egyptian are close variants of one another and hence they share considerable amount of lexical items. Some of the challenges faced by the participants include words that still have punctuation attached to them (e.g. \textRL{ (مصر}, (mSr, (Egypt ) . In order to mitigate these issues, some participants preprocessed these cases by, for example, removing any leading and trailing punctuation from those tokens. Other participants normalized these cases by unifying all the attached punctuations, while the remaining participants decided to keep them and let their model learn them. Table \ref{t:challenging_Arabic_test} and the following examples show some challenges faced by the submitted systems:
\begin{itemize}

\item Clitic attachment can obscure tokens, e.g. \textRL{والله}  wAllh “and-God” or "swear".

\item Clitic attachment can obscure tokens, e.g. \textRL{ومنى}  wmnY “and-Mona” or "swear". 
\end{itemize}



\begin{table}
\renewcommand{\arraystretch}{1.1}
\begin{center}
\begin{tabular}{|l|l|} \hline 
\bf N & \bf MSA-EGY Samples \\ \hline
1 & \underline{\textbf{Buckwalter Encoding:}}[\textbf{\textit{wAllh}}]\textsubscript{PER}  \\
  & OnA HAss bqhr In [\textbf{\textit{ElA' Ebd AlftAH}}]\textsubscript{PER}  \\
  & [\textbf{\textit{wmnY}}]\textsubscript{PER}  [\textbf{\textit{syf}}]\textsubscript{PER} bytHAkmwA  \\
  & wfy AlqfS \\
  & \underline{\textbf{Arabic:}}\textRL{والله أنا حاسس بقهر إن علاء } \\
  & \textRL{عبد الفتاح ومنى سيف}\\
  & \textRL{بيتحاكموا وفي القفص}\\
  &	\underline{\textbf{English:}} I swear I feel angry \\
  &  knowing that Ala Abdulfatah \\
  &  and-Mona are tried and jailed \\\hline

2 & \underline{\textbf{Buckwalter Encoding: }} kl wAHd  \\
  & ysOl Al|n :[\textbf{\textit{(mSr}}]\textsubscript{LOC} \\
  & rAyHp Ely fyn ?) \\
  &  \underline{\textbf{Arabic:}}\textRL{كل واحد يسأل الآن : } \\
  &  \textRL{(مصر رايحة علي فين ؟) } \\
  &	\underline{\textbf{English:}} Everyone asks himself   \\
  & where is Egypt going to go?  \\\hline
\end{tabular}
\end{center}
\caption{Challenging samples from the MSA-EGY test set. The bold words are the ground truth samples.}
\label{t:challenging_Arabic_test}
\end{table}

\section{Related work}

Before the CALCS workshop series, the code-switching behavior was studied from different perspectives and for many languages \citep{doi:10.1177/13670069010050040201, solorio-liu:2008:EMNLP1, Solorio:2008:PTE:1613715.1613852, piergallini-EtAl:2016:W16-58, DBLP:conf/acl-codeswitch/AlGhamdiMDSHSH16}. Most of them focused on either exploring this phenomenon or solving core code-switching tasks from the NLP pipeline. More recently, researchers have been considering the sentiment analysis task on code-switching settings \citep{lee-wang:2015:SIGHAN-8, W15-2902}. However, the lack of resources at the core level of the NLP pipeline greatly reduces the chances of improving higher-level applications. In this line, we aim at providing two datasets for named entity recognition benchmarks on the English-Spanish and Modern Standard Arabic-Egyptian language pairs.

It worth noting that there are some contributions of CS corpora, such as a collection of Turkish-German CS tweets \cite{DBLP:conf/lrec/2016}, a large collection of Modern Standrd Arabic and Egyptian Dialectal Arabic CS data \cite{DBLP:conf/lrec/DiabGHAAA16} and a collection of sentiment annotated Spanish-English tweets \cite{VILARES16.43}.
Named entity recognition has been vastly studied along the years \citep{DBLP:conf/conll/SangM03}. More recently, however, the focus has drastically moved to social media data due to the great incidence that social networks have in our daily communication \citep{Ritter:2011:NER:2145432.2145595, AugensteinEtAl:17}. The workshop on Noisy User-generated Text (W-NUT) has been a great effort towards the study of named entity recognition on noisy data. In 2016, the organizers focused on named entities from different topics to evaluate the adaptation of models from one topic to another \citep{StraussEtAl:16}. In 2017, the organizers introduced the Surface Form F1-score metric and collected data from multiple social media platforms \citep{DBLP:journals/corr/DerczynskiM0EGTPB14}. The challenge not only lies on the entity types and the social media noisy but also in the distribution of the datasets and their different data domain patterns.

\section{Conclusion}

We presented the setup and results of the 3rd shared task of the Computational Approaches to Linguistic Code-Switching workshop. We introduced a named entity recognition dataset focused on code-switched social media text for two language pairs: English-Spanish and Modern Standard Arabic-Egyptian. We received submissions from nine teams, eight of them submitted to ENG-SPA and six to MSA-EGY. Similar to the previous sequence tagging tasks of our workshop, the predominant aspect among the approaches was the Conditional Random Fields. Additionally, the combination of the CRF with a bidirectional LSTM (with some variations) yielded the best results among participants. The best F1-score for ENG-SPA was 63.7628\% and for MSA-EGY was 71.6154\%. Compared to monolingual formal text (i.e., newswire), the reported scores are significantly lower due to the code-switching phenomenon as well as the noise of SM environment. This serves as strong evidence that we need more robust approaches that can detect and process named entities in such challenging conditions.

\section*{Acknowledgments}
We would like to thank the National Science Foundation for partially supporting this work under award number 1462142.

\bibliography{acl2018}
\bibliographystyle{acl_natbib}

\end{document}